%% file: main.tex
\documentclass[letterpaper, 10 pt, conference]{cls/ieeeconf}  

\IEEEoverridecommandlockouts                              

\let\labelindent\relax

\usepackage{times}

\usepackage[table,usenames,dvipsnames]{xcolor}
\usepackage{enumitem}
\usepackage[noadjust]{cite}
\usepackage{amsmath,amssymb,amsfonts,amsthm,amscd,dsfont} 
\usepackage{algorithm,algorithmicx,listings}        
\usepackage[noend]{algpseudocode}			        

\usepackage{graphicx,tabularx,adjustbox}
\usepackage{multicol}
\usepackage[font={small}]{caption}   
\usepackage{subcaption}

\usepackage[bookmarks=true, breaklinks=true, colorlinks, citecolor=Black, urlcolor=Violet,linkcolor=Black]{hyperref}

\newtheorem{theorem}{Theorem}

\newtheorem*{proposition*}{Proposition}

\newtheorem*{corollary*}{Corollary}

\theoremstyle{definition}

\newtheorem*{assumption*}{Assumption}
\newtheorem*{problem*}{Problem}
\newtheorem{problem}{Problem}
\theoremstyle{remark}

\newtheorem*{solution*}{Solution}

\newcommand{\prl}[1]{\left(#1\right)}

\newcommand{\crl}[1]{\left\{#1\right\}}
\DeclareMathOperator*{\tr}{tr}
\DeclareMathOperator*{\diag}{diag}
\newcommand{\scaleMathLine}[2][1]{\resizebox{#1\linewidth}{!}{$\displaystyle{#2}$}}

\input{cls/sym.tex}

\title{\LARGE \bf
Adaptive Control of SE(3) Hamiltonian Dynamics with Learned Disturbance Features
}
\author{Thai Duong \and Nikolay Atanasov
\thanks{We gratefully acknowledge support from NSF RI IIS-2007141 and NSF CCF-2112665 (TILOS).}%
\thanks{The authors are with the Department of Electrical and Computer Engineering, University of California San Diego, La Jolla, CA 92093, USA, e-mail: {\tt\small \{tduong,natanasov\}@ucsd.edu}.}%
}

\begin{document}

\maketitle
\thispagestyle{empty}
\pagestyle{empty}

\begin{abstract}
Adaptive control is a critical component of reliable robot autonomy in rapidly changing operational conditions. Adaptive control designs benefit from a disturbance model, which is often unavailable in practice. This motivates the use of machine learning techniques to learn disturbance features from training data offline, which can subsequently be employed to compensate the disturbances online. This paper develops geometric adaptive control with a learned disturbance model for rigid-body systems, such as ground, aerial, and underwater vehicles, that satisfy Hamilton's equations of motion over the $SE(3)$ manifold. Our design consists of an \emph{offline disturbance model identification stage}, using a Hamiltonian-based neural ordinary differential equation (ODE) network trained from state-control trajectory data, and an \emph{online adaptive control stage}, estimating and compensating the disturbances based on geometric tracking errors. We demonstrate our adaptive geometric controller in trajectory tracking simulations of fully-actuated pendulum and under-actuated quadrotor systems.
\end{abstract}


\input{tex/Introduction.tex}

\input{tex/ProblemStatement.tex}
\input{tex/TechnicalApproach.tex}

\input{tex/Experiments.tex}

\input{tex/Conclusion.tex}
%

\bibliographystyle{cls/IEEEtran}
\bibliography{bib/thai_ref.bib}

\end{document}

%% file: cls/sym.tex
\newcommand{\calD}{{\cal D}}

\newcommand{\calH}{{\cal H}}

\newcommand{\calL}{{\cal L}}

\newcommand{\calR}{{\cal R}}

\newcommand{\calT}{{\cal T}}

\newcommand{\calV}{{\cal V}}

\newcommand{\frakp}{{\mathfrak{p}}}
\newcommand{\frakq}{{\mathfrak{q}}}

\newcommand{\bfa}{\mathbf{a}}

\newcommand{\bfd}{\mathbf{d}}
\newcommand{\bfe}{\mathbf{e}}
\newcommand{\bff}{\mathbf{f}}
\newcommand{\bfg}{\mathbf{g}}

\newcommand{\bfp}{\mathbf{p}}

\newcommand{\bfr}{\mathbf{r}}
\newcommand{\bfs}{\mathbf{s}}

\newcommand{\bfu}{\mathbf{u}}
\newcommand{\bfv}{\mathbf{v}}

\newcommand{\bfx}{\mathbf{x}}
\newcommand{\bfy}{\mathbf{y}}
\newcommand{\bfz}{\mathbf{z}}

\newcommand{\bfzeta}{{\boldsymbol{\zeta}}}

\newcommand{\bftheta}{{\boldsymbol{\theta}}}

\newcommand{\bfpi}{{\boldsymbol{\pi}}}
\newcommand{\bfrho}{{\boldsymbol{\rho}}}

\newcommand{\bftau}{{\boldsymbol{\tau}}}

\newcommand{\bfpsi}{{\boldsymbol{\psi}}}
\newcommand{\bfomega}{{\boldsymbol{\omega}}}

\newcommand{\bfE}{\mathbf{E}}

\newcommand{\bfI}{\mathbf{I}}
\newcommand{\bfJ}{\mathbf{J}}
\newcommand{\bfK}{\mathbf{K}}

\newcommand{\bfM}{\mathbf{M}}

\newcommand{\bfQ}{\mathbf{Q}}
\newcommand{\bfR}{\mathbf{R}}

\newcommand{\bfW}{\mathbf{W}}

\newcommand{\bfPsi}{\boldsymbol{\Psi}}

\newcommand{\bbR}{\mathbb{R}}

%% file: tex/Introduction.tex
\section{Introduction}
\label{sec:intro}

Autonomous mobile robots assisting in transportation, search and rescue, and environmental monitoring applications face complex and dynamic operational conditions. Ensuring safe operation depends on the availability of accurate system dynamics models, which can be obtained using system identification \cite{ljung1999system} or machine learning techniques \cite{nguyen2011model, deisenroth2015gp, greydanus2019hamiltonian, duong21hamiltonian}. 
When disturbances and system changes during online operation bring about new out-of-distribution data, it is often too slow to re-train the nominal dynamics model to support real-time adaptation to environment changes. Instead, adaptive control \cite{krstic1995nonlinear, ioannou1996robust} offers efficient tools to estimate and compensate for disturbances and parameter variations online. 

A key technical challenge in adaptive control is the parameterization of the system uncertainties \cite{krstic1995nonlinear}. For linearly parameterized systems, disturbances may be modeled as linear combinations of \emph{known} nonlinear features and updated by an adaptation law based on the state errors with stability obtained by sliding-mode theory \cite{slotine1987adaptive, slotine1990hamiltonian, dirksz2012structure}, assuming zero-state detectability \cite{sastry1989adaptive, dirksz2012structure} or $\calL_1$-adaptation \cite{hovakimyan2010L1, gahlawat20al1adaptive,hanover2021performance}. If the system evolves on a manifold (e.g., when the state contains orientation), an adaptation law is designed based on geometric errors, derived from the manifold constraints \cite{goodarzi2015geometric, bisheban2020geometric}. A disturbance observer \cite{chen2000nonlinear, li2014disturbance} use the state errors introduced by the disturbances to design an asymptotically stable observer system that estimates the disturbance online. A disturbance adaptation law is paired with a nominal controller, derived using Lagrangian dynamics with feedback linearization \cite{slotine1987adaptive, slotine1991applied}, Hamiltonian dynamics with energy shaping \cite{nageshrao2015port, dirksz2012structure}, or model predictive control \cite{pereida2018adaptive,hanover2021performance}.

Recently, there has been growing interest in applying machine learning techniques to design adaptive controllers. As the nonlinear disturbance features are actually \emph{unknown} in practice, they can be estimated using Gaussian processes \cite{gahlawat20al1adaptive, grande2013nonparametric} or neural networks \cite{joshi2020asynchronous,  richards21adaptive}. The features can be learned online in the control loop \cite{gahlawat20al1adaptive, joshi2020asynchronous}, which is potentially slow for real-time operation, or offline via meta-learning from past state-control trajectories \cite{harrison2018meta} or system dynamics simulation \cite{richards21adaptive}. Given the learned disturbance features, an adaptation law is designed to estimate the disturbances online, e.g. using $\calL_1$-adaptation \cite{gahlawat20al1adaptive} or by updating the last layer of the feature neural network \cite{joshi2020asynchronous, richards21adaptive, shi2021meta}.

This paper develops data-driven adaptive control for rigid-body systems, such as unmanned ground vehicles (UGVs), unmanned aerial vehicles (UAVs), or unmanned underwater vehicles (UUVs), that satisfy Hamilton's equations of motion on position and orientation manifold $SE(3)$. While recent techniques for disturbance feature learning and data-driven adaptive control are restricted to systems whose states evolve in Euclidean space, a unique aspect of our adaptive control design is the consideration of geometric tracking errors on the $SE(3)$ manifold. Compared to existing $SE(3)$ geometric adaptive controllers specifically designed for quadrotors with a known disturbance model \cite{goodarzi2015geometric, bisheban2020geometric}, we develop a general adaptation law that can be used for any rigid-body robot, such as a UGV, UAV, or UUV, and learn disturbance features from trajectory data instead of assuming a known model. Specifically, given a dataset of state-control trajectories with different disturbance realizations, we learn nonlinear disturbance features using a Hamiltonian-based neural ODE network \cite{chen2018neural, duong21hamiltonian}, where the disturbances are represented by a neural network, connected in an architecture that respects the Hamiltonian dynamics. We develop a geometric adaptation law to estimate the disturbances online and compensate them by a nonlinear energy-shaping tracking controller. 

In summary, our \textbf{contribution} is a learning-based adaptive geometric control for $SE(3)$ Hamiltonian dynamics that
\begin{itemize}
	\item learns disturbance features offline from state-control trajectories using an $SE(3)$ Hamiltonian-based neural ODE network, and
	\item employs energy-based tracking control with adaptive disturbance compensation online based on the learned disturbance model and the geometric tracking errors.
\end{itemize}
We verify our approach using simulated fully-actuated pendulum and under-actuated quadrotor systems, and compare with a disturbance observer method to highlight the benefit of learning disturbance features from data.

%% file: tex/ProblemStatement.tex
\section{Problem Statement}
\label{sec:problem_statement}

Consider a system modeled as a single rigid body with position $\bfp \in \bbR^3$, orientation $\bfR\in SO(3)$, body-frame linear velocity $\bfv \in \bbR^3$, and body-frame angular velocity $\bfomega \in \bbR^3$. Let $\frakq = [\bfp^\top\;\; \bfr_1^\top\;\; \bfr_2^\top\;\; \bfr_3^\top]^\top \in SE(3)$ be the generalized coordinates, where $\bfr_1$, $\bfr_2$, $\bfr_3$ are the rows of the rotation matrix $\bfR$. Let $\bfzeta = [\bfv^\top\;\bfomega^\top]^\top \in \bbR^6$ be the generalized velocity. The generalized momentum of the system is defined as $\frakp = \bfM(\frakq)\bfzeta \in \bbR^6$,
%
%
where $\bfM(\frakq) \in \bbR^{6\times6}$ is the generalized mass matrix. The state is defined as $\bfx = (\frakq, \frakp)$ and its evolution is governed by the system dynamics:
\begin{equation} \label{eq:dynamics}
\dot{\bfx} = \bff(\bfx, \bfu, \bfd),
\end{equation}
where $\bfu$ is the control input and $\bfd$ is a disturbance signal. The disturbance $\bfd$ is modeled as a linear combination of nonlinear features $\bfW(\bfx)\in \bbR^{6\times p}$:
\begin{equation}
\label{eq:f_external}
\bfd(t) = \bfW(\bfx(t))\bfa^*,
\end{equation}
where $\bfa^* \in \bbR^p$ are unknown feature weights.

A mechanical system obeys Hamilton's equations of motion \cite{lee2017global}. The Hamiltonian, $\mathcal{H}(\mathbf\frakq, \mathbf\frakp) = T(\mathbf\frakq, \mathbf\frakp) + V(\mathbf\frakq)$, captures the total energy of the system as the sum of the kinetic energy $T(\mathbf\frakq, \mathbf\frakp) =  \frac{1}{2}\mathbf\frakp^\top \bfM(\mathbf\frakq)^{-1} \mathbf\frakp$ and the potential energy $V(\frakq)$.
%
%
The dynamics in \eqref{eq:dynamics} are determined by the Hamiltonian \cite{lee2017global, forni2015port} and have a port-Hamiltonian structure \cite{van2014port, duong21hamiltonian}:
%
\begin{equation}
\label{eq:PH_dyn}
\begin{bmatrix}
\dot{\mathbf\frakq} \\
\dot{\mathbf\frakp} \\
\end{bmatrix}
= \begin{bmatrix}
\bf0 & \mathbf\frakq^{\times} \\
-\mathbf\frakq^{\times\top} & \mathbf\frakp^{\times} 
\end{bmatrix}
\begin{bmatrix}
\frac{\partial \mathcal{H}}{\partial\mathbf\frakq} \\
\frac{\partial \mathcal{H}}{\partial\mathbf\frakp} 
\end{bmatrix} + \begin{bmatrix} \bf0 \\ \bfg(\mathbf\frakq) \end{bmatrix}\bfu + \begin{bmatrix} \bf0 \\ \bfd \end{bmatrix},
\end{equation}
where $\bfg(\mathbf\frakq)$ is the input gain and the disturbance $\bfd$ appears as an external force applied to the system. 
The operators $\mathbf\frakq^{\times}$ and $\mathbf\frakp^{\times}$ are defined as:
\begin{equation*}
\mathbf\frakq^{\times} = \begin{bmatrix}
\bfR^\top\!\!\!\! & \bf0 & \bf0 & \bf0 \\
\bf0 & \hat{\bfr}_1^\top & \hat{\bfr}_2^\top & \hat{\bfr}_3^\top
\end{bmatrix}^\top\!\!\!\!, \quad \mathbf\frakp^{\times} = \begin{bmatrix} \mathbf\frakp_{\bfv}\\\mathbf\frakp_{\bfomega}\end{bmatrix}^{\times} \!\!\!\!= \begin{bmatrix}
\bf0 & \hat{\mathbf\frakp}_{\bfv}\\
\hat{\mathbf\frakp}_{\bfv} & \hat{\mathbf\frakp}_{\bfomega}
\end{bmatrix},
\end{equation*}
where the hat map $\hat{(\cdot)}: \bbR^3 \mapsto \mathfrak{so}(3)$ constructs a skew-symmetric matrix from a 3D vector. Note that the equation $\dot{\frakq} = \frakq^\times \frac{\partial \calH}{\partial\frakq}$ in \eqref{eq:PH_dyn} exactly specifies the $SE(3)$ kinematics, $\dot{\bfp} = \bfR\bfv$ and $\dot{\bfR} = \bfR \hat{\bfomega}$, with the rotation part written row-by-row.

Consider a collection $\calD = \{\calD_1, \calD_2, \ldots, \calD_M\}$ of system state transitions $\calD_j$, each obtained under a different unknown disturbance realization $\bfa_j^*$, for $j = 1,\ldots,M$. Each $\calD_j = \{\bfx^{(ij)}_{0}, \bfu^{(ij)}, \bfx^{(ij)}_{f}, \tau^{(ij)}\}_{i = 1}^{D_j}$ consists of $D_j$ state transitions, each obtained by applying a constant control input $\bfu^{(ij)}$ to the system with initial condition $\bfx^{(ij)}_0$ and sampling the state $\bfx^{(ij)}_f := \bfx^{(ij)}(\tau^{(ij)})$ at time $\tau^{(ij)}$. Our objective is to approximate the disturbance model in \eqref{eq:f_external} by $\bar{\bfd}_\bftheta(t) = \bfW_\bftheta(\bfx(t)) \bfa_j$, where $\bftheta$ parameterizes the shared disturbance features and the parameters $\{\bfa_j\}_{j=1}^M$ model each disturbance realization. To optimize $\bftheta$, $\{\bfa_j\}$, we predict the dynamics evolution starting from state $\bfx^{(ij)}_0$ with control $\bfu^{(ij)}$ and minimize the distance between the predicted state $\bar{\bfx}^{(ij)}_{f}$ and the true state $\bfx^{(ij)}_{f}$ from $\calD_j$, for $j = 1, \ldots, M$. Since the approximated disturbance $\bar{\bfd}_\bftheta$ does not change if the features $\bfW_{\bftheta}$ and the coefficients $\bfa_j$ are scaled by constants $\gamma$ and $1/\gamma$, respectively, we add the norms of  $\bfW_{\bftheta}(\bfx^{(ij)}_0)$ and $\{\bfa_j\}_{j = 1}^M$ to the objective function as regularization terms.

\begin{problem}
	\label{problem:dynamics_learning}
	Given $\calD = \{\{\bfx^{(ij)}_{0}, \bfu^{(ij)}, \bfx^{(ij)}_{f}, \tau^{(ij)}\}_{i = 1}^{D_j}\}_{j=1}^M$, find disturbance parameters $\bftheta$, $\{\bfa_j\}_{j=1}^M$ that minimize:
	\begin{align} \label{problem_formulation_unknown_env_equation}
	\min_{\bftheta, \{\bfa_j\}} \;&\sum_{j=1}^M \sum_{i = 1}^{D_j}  \ell(\bfx^{(ij)}_f,\bar{\bfx}^{(ij)}_f) + \notag\\
	&\qquad \lambda_\bftheta \sum_{j=1}^M \sum_{i = 1}^{D_j} \Vert \bfW_{\bftheta}(\bfx^{(ij)}_0)\Vert^2  + \lambda_\bfa\sum_{j=1}^M \Vert \bfa_j \Vert^2 \notag\\
	\text{s.t.} \;\; & \dot{\bar{\bfx}}^{(ij)}(t) = {\bff}(\bar{\bfx}^{(ij)}(t), \bfu^{(ij)}, \bar{\bfd}_{\bftheta}^{(ij)}(t)),\\
	& \bar{\bfd}_{\bftheta}^{(ij)}(t) = \bfW_{\bftheta}(\bar{\bfx}^{(ij)}(t)) \bfa_j,\notag\\
	& \;\;\bar{\bfx}^{(ij)}(0) = \bfx^{(ij)}_0,\;\;\bar{\bfx}^{(ij)}_f = \bar{\bfx}^{(ij)}(\tau^{(ij)}), \notag\\
	&\forall i = 1, \ldots, D_j, \;\;\forall j = 1, \ldots, M,\notag
	\end{align}
	where $\ell$ is a distance metric on the state space.
\end{problem}

After the offline disturbance feature identification in Problem~\ref{problem:dynamics_learning}, we design a controller $\bfu = \bfpi(\bfx, \bfx^*, \bfa; \bftheta)$ that tracks a desired state trajectory $\bfx^*(t)$, using the dynamics ${\bff}$ and the learned disturbance model $\bfW_\bftheta(\bfx)$. To handle a disturbance signal $\bfd(t) = \bfW_\bftheta(\bfx(t))\bfa^*$ with an unknown realization $\bfa^*$, we augment the tracking controller with an adaptation law $\dot{\bfa} = \bfrho(\bfx, \bfx^*, \bfa; \bftheta)$, estimating $\bfa^*$ online, so that $\limsup_{t \to \infty} \ell(\bfx(t),\bfx^*(t))$ is bounded.

%% file: tex/TechnicalApproach.tex
\section{Technical Approach}
\label{sec:technical_approach}

We present our approach in two stages: disturbance feature learning to solve Problem \ref{problem:dynamics_learning} (Sec. \ref{subsec:Ham_dyn_learning}) and geometric adaptive control design for trajectory tracking (Sec. \ref{subsec:adaptive_control_online}).

\subsection{$SE(3)$ Hamiltonian-based disturbance feature  learning}
\label{subsec:Ham_dyn_learning}

To address Problem~\ref{problem:dynamics_learning}, we use a neural ODE network \cite{chen2018neural} whose structure respects Hamilton's equations in \eqref{eq:PH_dyn} with known generalized mass $\bfM(\frakq)$, potential energy $V(\frakq)$ and the input gain $\bfg(\frakq)$. We introduce a disturbance model, $\bfd = \bfW_\bftheta(\frakq, \frakp) \bfa$, where $\bfW_\bftheta(\frakq, \frakp)$ is a neural network, and estimate its parameters $\bftheta$ from disturbance-corrupted data. The training data $\calD_j = \{\bfx^{(ij)}_{0}, \bfu^{(ij)}, \bfx^{(ij)}_{f}, \tau^{(ij)}\}_{i = 1}^{D_j}$ may be obtained using an odometry algorithm \cite{OdometrySurvey} or a motion capture system. The data collection can be performed using an existing baseline controller or a human operator manually controlling the system under different disturbance conditions (e.g., wind, ground effect, etc. for a UAV).

We define the geometric distance metric $\ell$ in Problem~\ref{problem:dynamics_learning} as a sum of position, orientation, and momentum errors:
\begin{equation}
\ell(\bfx,\bar{\bfx}) = \ell_{\bfp}(\bfx,\bar{\bfx}) + \ell_{\bfR}(\bfx,\bar{\bfx}) + \ell_{\frakp}(\bfx,\bar{\bfx}),
\end{equation}
where $\ell_{\bfp}(\bfx,\bar{\bfx}) = \| \bfp - \bar{\bfp}\|^2_2$, $\ell_{\frakp}(\bfx,\bar{\bfx})~=~\| \frakp~-~\bar{\frakp}\|^2_2$,
$\ell_{\bfR}(\bfx,\bar{\bfx}) = \|\left(\log (\bar{\bfR} \bfR^\top)\right)^{\vee} \|_2^2$,
%
%
$\log : SE(3) \mapsto \mathfrak{so}(3)$ is the inverse of the exponential map, associating a rotation matrix to a skew-symmetric matrix, and $(\cdot)^\vee : \mathfrak{so}(3) \mapsto \bbR^3$ is the inverse of the hat map $\hat{(\cdot)}$.
Let $\calL(\bftheta, \{\bfa_j\}; \calD)$ be the total loss in Problem~\ref{problem:dynamics_learning}. To calculate the loss, for each dataset $\calD_j$ with disturbance $\bar{\bfd}_\bftheta^{(ij)}(t) = \bfW_\bftheta(\bar{\bfx}^{(ij)}(t)) \bfa_j$, we solve an ODE:
\begin{equation}
\dot{\bar{\bfx}}^{(ij)} = {\bff}(\bar{\bfx}^{(ij)}, \bfu^{(ij)}, \bar{\bfd}_\bftheta^{(ij)}), \quad \bar{\bfx}^{(ij)}(0) = \bfx^{(ij)}_{0},
\end{equation}
using an ODE solver. This generates a predicted state $\bar{\bfx}^{(ij)}_{f}$ at time $\tau^{(ij)}$ for each $i = 1,\ldots, D_j$ and $j = 1, \ldots, M$:
\begin{equation}
{\bar{\bfx}}_f^{(ij)} = \text{ODESolver}\prl{\bfx_0^{(ij)}, {\bff}, \tau^{(ij)}; \bftheta},
\end{equation}
sufficient to compute $\calL(\bftheta, \{\bfa_j\}; \calD)$. The parameters $\bftheta$ and $\{\bfa_j\}$ are updated using gradient descent by back-propagating the loss through the neural ODE solver using adjoint states $\bfy = \frac{\partial \calL}{\partial \bar{\bfx}}$ \cite{chen2018neural}. An augmented state $\bfs = \prl{\bar{\bfx}, \bfy, \frac{\partial \calL}{\partial \bftheta}, \crl{\frac{\partial \calL}{\partial \bfa_j}}}$ satisfies $\dot{\bfs} = \bff_\bfs = \prl{\bff, -\bfy^\top \frac{\partial \bff}{\partial \bar{\bfx}}, -\bfy^\top \frac{\partial \bff}{\partial \bftheta}, \crl{-\bfy^\top \frac{\partial \bff}{\partial \bfa_j}}}$. The gradients $\frac{\partial \calL}{\partial \bftheta}$  and $\crl{\frac{\partial \calL}{\partial \bfa_j}}$ are obtained by a call to a reverse-time ODE solver starting from $\bfs_f = \bfs_f(\tau^{(ij)})$:
\begin{equation}
\scaleMathLine[0.9]{\bfs_0 = \prl{\bar{\bfx}_0, \bfa_0, \frac{\partial \calL}{\partial\bftheta}, \crl{\frac{\partial \calL}{\partial \bfa_j}}} = \text{ODESolver}(\bfs_f, \bff_s, \tau^{(ij)})}.
\end{equation}
%



\subsection{Data-driven geometric adaptive control}
\label{subsec:adaptive_control_online}

Given the learned disturbance model $\bfW_\bftheta(\bfx)$ and a desired trajectory $\bfx^*(t)$, we develop a trajectory tracking controller $\bfu = \bfpi(\bfx, \bfx^*, \bfa; \bftheta)$ that compensates for disturbances and an adaptation law $\dot{\bfa} = \bfrho(\bfx, \bfx^*, \bfa; \bftheta)$ that estimates the disturbance realization online.

Our tracking controller for the Hamiltonian dynamics in \eqref{eq:PH_dyn} is developed using interconnection and damping assignment passivity-based control (IDA-PBC) \cite{van2014port}. Consider a desired pose-velocity trajectory $(\mathbf\frakq^*(t), \bfzeta^*(t))$. Since the momentum $\frakp$ is defined in the body inertial frame, the desired momentum $\frakp^*(t)$ should be computed by transforming the desired velocity $\bfzeta^* = [\bfv^{*\top}\; \bfomega^{*\top}]^\top$ to the body frame as $\mathbf\frakp^* = \bfM (\frakq) \begin{bmatrix} \bfR^\top \bfR^* \bfv^* \\ \bfR^\top \bfR^* \bfomega^* \end{bmatrix}$.
%
%
The Hamiltonian of the system \eqref{eq:PH_dyn} is not necessarily minimized along $\bfx^*(t) = (\mathbf\frakq^*(t), \mathbf\frakp^*(t))$. The key idea of an IDA-PBC design is to choose the control input $\bfu(t)$ so that the closed-loop system has a desired Hamiltonian $\calH_d(\mathbf\frakq, \mathbf\frakp)$, which is minimized along $\bfx^*(t)$. Using quadratic errors in the position, orientation, and momentum, we design the desired Hamiltonian:
\begin{align} \label{eq:desired_hamiltonian}
\mathcal{H}_d&(\mathbf\frakq, \mathbf\frakp) =  \frac{1}{2}k_\bfp(\bfp - \bfp^*)^\top(\bfp - \bfp^*) \\
& + \frac{1}{2} k_{\bfR}\tr(\bfI - \bfR^{*\top}\bfR) + \frac{1}{2}(\mathbf\frakp-\mathbf\frakp^*)^\top\bfM^{-1}(\mathbf\frakq)(\mathbf\frakp-\mathbf\frakp^*),\notag
\end{align}
where $k_\bfp$ and $k_{\bfR}$ are positive gains. We solve a set of matching conditions, described in \cite{duong21hamiltonian, van2014port}, between the original dynamics \eqref{eq:PH_dyn} with Hamiltonian $\calH(\mathbf\frakq, \mathbf\frakp)$ and the desired dynamics with Hamiltonian $\calH_d(\mathbf\frakq, \mathbf\frakp)$ in \eqref{eq:desired_hamiltonian} to arrive at a tracking controller $\bfu = \bfpi(\bfx,\bfx^*,\bfa ; \bftheta)$. The controller consists of an energy-shaping term $\bfu_{ES}$, a damping-injection term $\bfu_{DI}$, and a disturbance compensation term $\bfu_{DC}$:
\begin{align}
\label{eq:ES_DI_COMP_control}
\bfu_{ES} &=\bfg^{\dagger}(\mathbf\frakq)\left(\mathbf\frakq^{\times\top} \frac{\partial V}{\partial \mathbf\frakq} - \mathbf\frakp^{\times}\bfM^{-1}(\mathbf\frakq)\mathbf\frakp - \bfe(\mathbf\frakq,\mathbf\frakq^*) + \dot{\frakp}^*\right), \notag\\
\bfu_{DI} &= -\bfK_\bfd\bfg^{\dagger}(\mathbf\frakq)   \bfM^{-1}(\mathbf\frakq)(\mathbf\frakp-\mathbf\frakp^*),\\
\bfu_{DC}&= -\bfg^{\dagger}(\mathbf\frakq)\bfW(\mathbf\frakq, \mathbf\frakp) \bfa, \notag
\end{align}
where $\bfg^{\dagger}(\mathbf\frakq) = \left(\bfg^{\top}(\mathbf\frakq)\bfg(\mathbf\frakq)\right)^{-1}\bfg^{\top}(\mathbf\frakq)$ is the pseudo-inverse of $\bfg(\mathbf\frakq)$ and $\bfK_\bfd = \diag(k_\bfv \bfI, k_\bfomega\bfI)$ is a damping gain with positive terms $k_\bfv$, $k_\bfomega$. The controller utilizes a generalized coordinate error between $\mathbf\frakq$ and $\mathbf\frakq^*$:
\begin{equation}
\label{eq:coordinate_error}
\scaleMathLine[0.89]{\bfe(\mathbf\frakq,\mathbf\frakq^*) = \begin{bmatrix} \bfe_{\bfp}(\mathbf\frakq,\mathbf\frakq^*) \\ \bfe_{\bfR}(\mathbf\frakq,\mathbf\frakq^*) \end{bmatrix} = \begin{bmatrix} k_\bfp \bfR^\top(\bfp - \bfp^*) \\ \frac{1}{2}k_{\bfR}\prl{\bfR^{*\top}\bfR-\bfR^\top\bfR^{*}}^{\vee}\end{bmatrix}}
\end{equation}
and a generalized momentum error $ \mathbf\frakp_e = \mathbf\frakp - \mathbf\frakp^*$:
\begin{equation}
\label{eq:momenta_error}
\scaleMathLine[0.89]{\mathbf\frakp_e = \bfM(\mathbf\frakq) \begin{bmatrix} \bfe_{\bfv}(\bfx,\bfx^*) \\ \bfe_{\bfomega}(\bfx,\bfx^*) \end{bmatrix}= \bfM(\mathbf\frakq) \begin{bmatrix} \bfv - \bfR^\top \bfR^* \bfv^* \\ \bfomega - \bfR^\top \bfR^* \bfomega^* \end{bmatrix}.}
\end{equation}
Please refer to \cite{duong21hamiltonian} for a detailed derivation of $\bfu_{ES}$ and $\bfu_{DI}$. 


The disturbance compensation term $\bfu_{DC}$ in \eqref{eq:ES_DI_COMP_control} requires online estimation of the disturbance feature weights $\bfa$. Inspired by \cite{goodarzi2015geometric}, we design an adaptation law which utilizes the geometric errors \eqref{eq:coordinate_error}, \eqref{eq:momenta_error} to update the weights $\bfa$:
\begin{equation}
\label{eq:geometric_adaptive_law}
\begin{aligned}
\dot{\bfa} &= \bfrho(\bfx,\bfx^*,\bfa; \bftheta)\\
& = \bfW_{\bftheta}^\top(\mathbf\frakq, \mathbf\frakp) \begin{bmatrix} c_{\bfp} \bfe_{\bfp}(\mathbf\frakq,\mathbf\frakq^*) + c_{\bfv} \bfe_{\bfv}(\bfx,\bfx^*) \\ c_{\bfR} \bfe_{\bfR}(\mathbf\frakq,\mathbf\frakq^*) + c_{\bfomega} \bfe_{\bfomega}(\bfx,\bfx^*) \end{bmatrix},
\end{aligned}
\end{equation}
where $c_{\bfp}$, $c_{\bfv}$, $c_{\bfR}$, $c_{\bfomega}$ are positive coefficients. The stability of our adaptive controller $(\bfpi, \bfrho)$ is shown in Theorem \ref{thm:stability}.

\begin{theorem} \label{thm:stability}
	Consider the Hamiltonian dynamics in \eqref{eq:PH_dyn} with disturbance model in \eqref{eq:f_external}. Suppose that the parameters $\bfg(\mathbf\frakq)$, $\bfM(\mathbf\frakq)$, $V(\mathbf\frakq)$, and $\bfW(\mathbf\frakq,\mathbf\frakp)$ are known but the distubance feature weights $\bfa^*$ are unknown. Let $\bfx^*(t)$ be a desired state trajectory with bounded angular velocity, $\|\bfomega^*(t)\| \leq \gamma$. Assume that the initial system state lies in the domain $\calT = \crl{ \bfx \in T^*SE(3) \mid \bfPsi(\bfR, \bfR^*) < \alpha < 2, \|\bfe_{\bfomega}(\bfx,\bfx^*)\| < \beta}$ for some positive constants $\alpha$ and $\beta$, where $\bfPsi (\bfR, \bfR^*) = \frac{1}{2}\tr(\bfI - \bfR^{*\top}\bfR)$. 
	Consider the tracking controller in \eqref{eq:ES_DI_COMP_control} with adaptation law in \eqref{eq:geometric_adaptive_law}. Then, there exist positive constants $k_{\bfp}$, $k_{\bfR}$, $k_{\bfv}$, $k_{\bfomega}$, $c_{\bfp} = c_{\bfR} = c_1$, $c_{\bfv} = c_{\bfomega}=c_2$ such that the tracking errors $\bfe(\mathbf\frakq,\mathbf\frakq^*)$ and $\mathbf\frakp_e$ defined in \eqref{eq:coordinate_error} and \eqref{eq:momenta_error} converge to zero. Also, the estimation error $\bfe_{\bfa} = \bfa - \bfa^*$ is stable in the sense of Lyapunov and uniformly bounded. An estimate of the region of attraction is $\calR = \{\bfx\in\calT\mid\calV(\bfx) \leq \delta\}$, where:
	\begin{equation}\label{eq:lyapunov-function}
	    \calV(\frakq, \frakp) = \calH_d(\frakq, \frakp) +\frac{c_1}{c_2}\bfe^\top \mathbf\frakp_e + \frac{1}{2c_2} \|\bfe_{\bfa}\|_2^2
	\end{equation}
	and $\delta < \lambda_{\min} (\bfQ_1)\min(\alpha (2-\alpha),\beta)/2$ for
	\begin{equation} \label{eq:Q1}
	\bfQ_1 = \begin{bmatrix} \min\crl{k_{\bfp}, k_{\bfR}} & - {c_1}/{c_2} \\
	- {c_1}/{c_2}  & \lambda_{\min}(\bfM^{-1}(\frakq)) \end{bmatrix}.  
	\end{equation}
\end{theorem}

\begin{figure*}[!t]
	\begin{subfigure}[t]{0.3\textwidth}
		\centering
		\includegraphics[width=\textwidth]{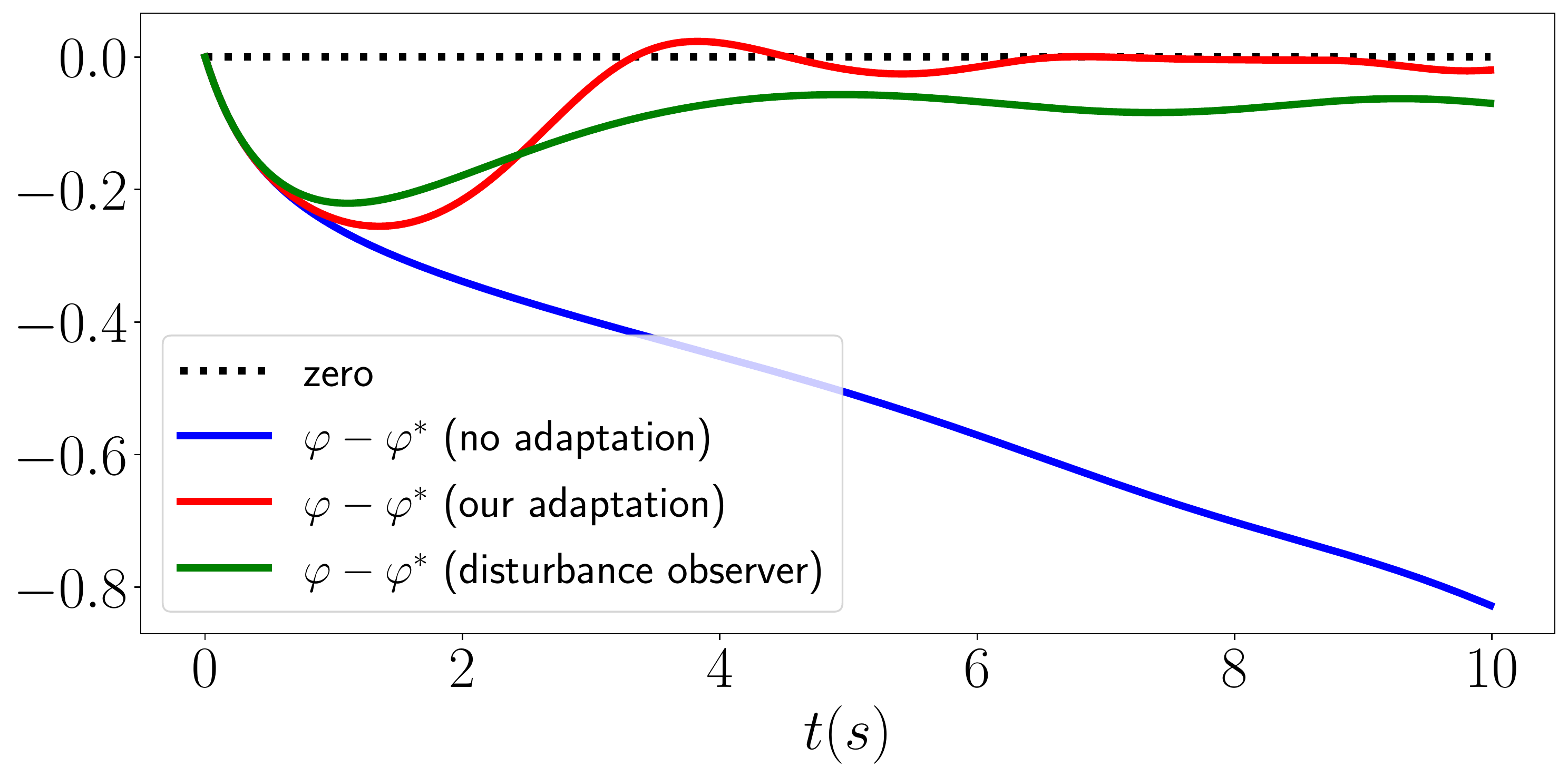}
		\caption{}
		\label{fig:pend_phi_error}
	\end{subfigure}%
	\hfill%
	\begin{subfigure}[t]{0.3\textwidth}
		\centering
		\includegraphics[width=\textwidth]{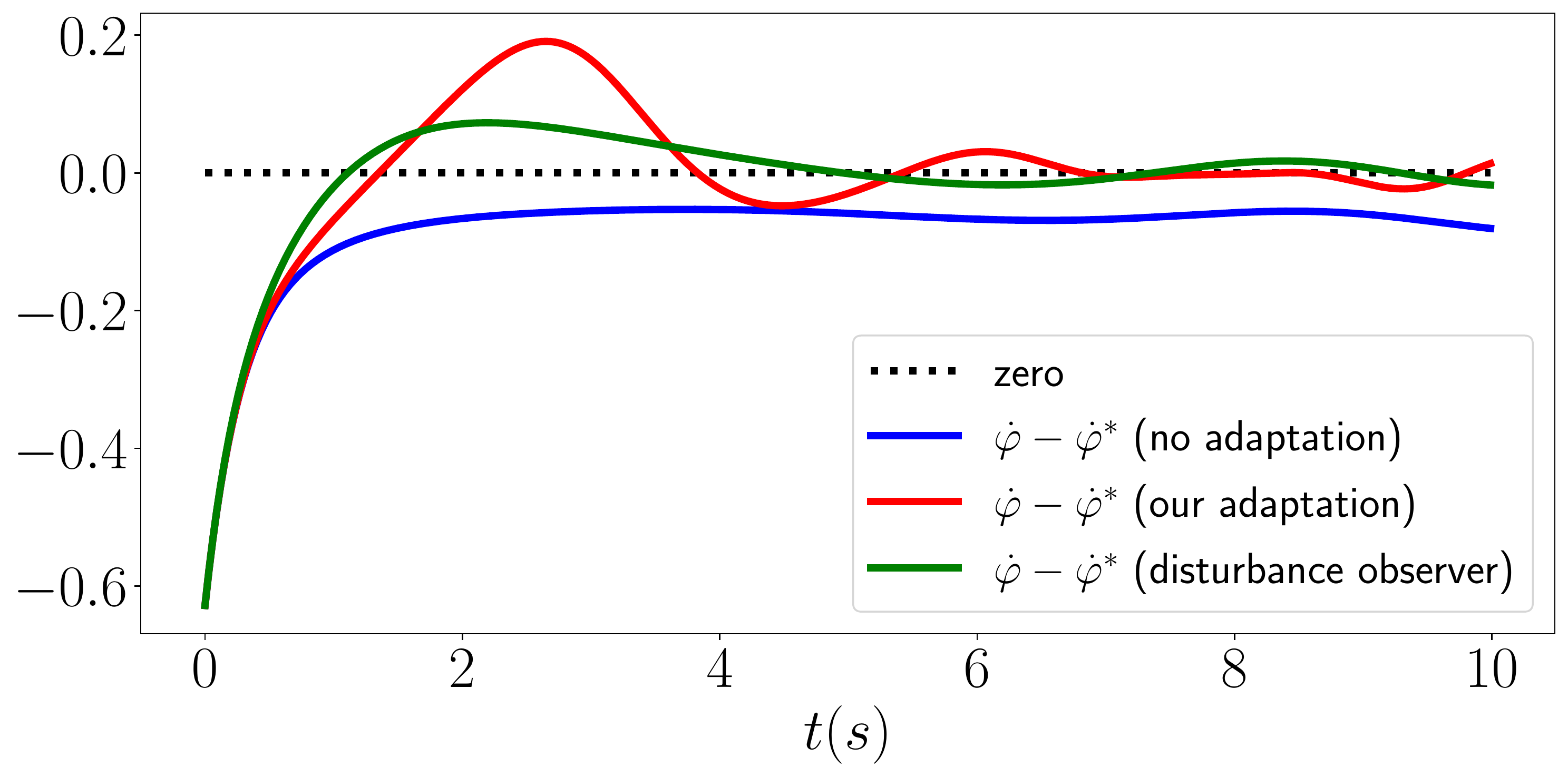}%
		\caption{}
		\label{fig:pend_phidot_error}
	\end{subfigure}%
	\hfill%
	\begin{subfigure}[t]{0.30\textwidth}
		\centering
		\includegraphics[width=\textwidth]{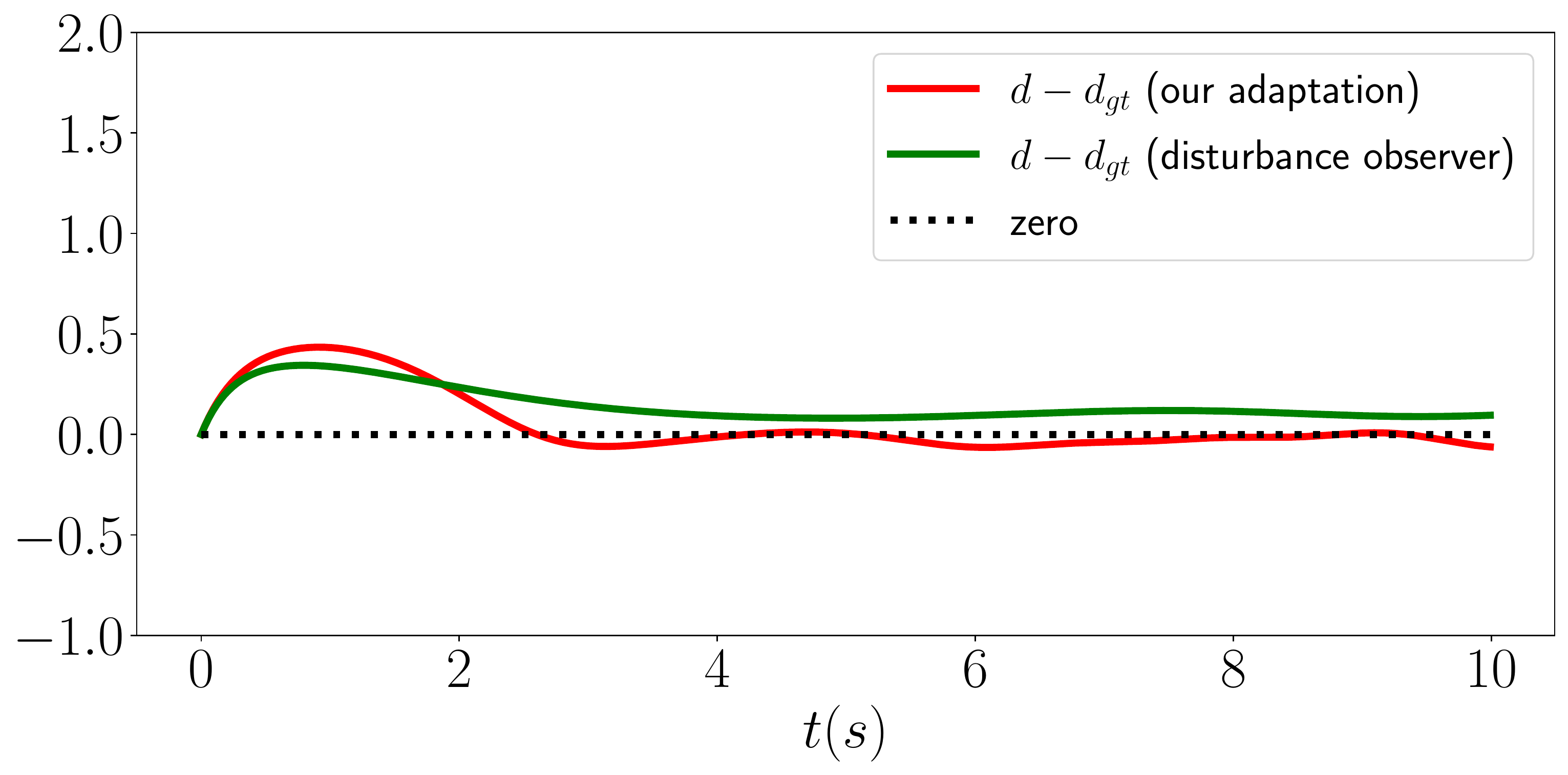}%
		\caption{}
		\label{fig:pend_disturbance_error}
	\end{subfigure}%
	\caption{Comparison of our learned adaptive controller and a disturbance observer method on a pendulum: (a) the angle error $\varphi -  \varphi^*$; (b) velocity error $\dot{\varphi} - \dot{\varphi^*}$; and (c) the disturbance error $d - d_{gt}$ with the ground-truth disturbance $d_{gt} = -2.5 \dot{\varphi}$.}
	\label{fig:pend_exp}
\end{figure*}

\begin{proof}
	We drop function parameters to simplify the notation. The derivative of the generalized coordinate error satisfies:
	\begin{equation}
	\begin{aligned} \label{eq:edot}
	\dot{\bfe} &=  \begin{bmatrix}
	\dot{\bfe}_{\bfp} \\ \dot{\bfe}_{\bfR}
	\end{bmatrix} = \begin{bmatrix} -\hat{\bfomega} \bfe_{\bfp} + k_{\bfp}\bfe_{\bfv}\\
	k_{\bfR}\bfE(\bfR,\bfR^*) \bfe_{\bfomega}
	\end{bmatrix}\\
	&= -\begin{bmatrix} \hat{\bfomega} & \mathbf{0}\\ \mathbf{0}& \mathbf{0} \end{bmatrix} \bfe + \begin{bmatrix} k_{\bfp} \bfI & \mathbf{0}\\\mathbf{0} & k_{\bfR}\bfE(\bfR,\bfR^*) \end{bmatrix} \bfM^{-1} \mathbf\frakp_e,
	\end{aligned}
	\end{equation}
	where $\bfE(\bfR,\bfR^*) = \frac{1}{2}\prl{\tr(\bfR^\top \bfR^*) \bfI - \bfR^\top \bfR^* }$ satisfies $\|\bfE(\bfR,\bfR^*)\| \leq 1$. By construction of the IDA-PBC controller \cite{duong21hamiltonian}:
	\begin{equation}\label{eq:frakpedot}
	\dot{\mathbf\frakp}_e = -\bfe - \bfK_d \bfM^{-1} \mathbf\frakp_e - \bfW\bfe_{\bfa}.
	\end{equation}
	Consider the adaptation law $\dot{\bfa} = c_1 \bfW^\top\bfe + c_2 \bfW^\top\bfM^{-1}\mathbf\frakp_e$ in \eqref{eq:geometric_adaptive_law} with $c_1 = c_{\bfp} = c_{\bfR}$ and $c_2 = c_{\bfv}=c_{\bfomega}$. In the domain $\calT$, $\bfPsi(\bfR, \bfR^*) < \alpha < 2$ and $\frac{1}{2}\|\bfe_{\bfR}\|_2^2 \leq \bfPsi(\bfR, \bfR^*) \leq \frac{1}{2-\alpha} \|\bfe_{\bfR}\|_2^2$ by \cite[Prop.~1]{goodarzi2015geometric}.
	For $\bfz~:=~[\|\bfe\|\; \|\mathbf\frakp_e\|]^\top \in \bbR^2$, the Lyapunov function candidate $\calV$ in \eqref{eq:lyapunov-function} is bounded as:
	\begin{equation}
	\frac{1}{2}\bfz^\top \bfQ_1 \bfz + \frac{1}{2c_2} \|\bfe_{\bfa}\|_2^2 \leq \calV \leq \frac{1}{2}\bfz^\top \bfQ_2 \bfz + \frac{1}{2c_2} \|\bfe_{\bfa}\|_2^2,
	\end{equation}
	where the matrix $\bfQ_1$ is specified in \eqref{eq:Q1} and $\bfQ_2$ is:
	\begin{equation}
	\bfQ_2 = \begin{bmatrix} \max\crl{k_{\bfp}, \frac{2 k_{\bfR}}{2 - \alpha}} &  {c_1}/{c_2} \\
	{c_1}/{c_2}  & \lambda_{\max}(\bfM^{-1}) \end{bmatrix}.
	\end{equation}
	%
	%
	%
	The time derivative of the Lyapunov candidate satisfies:
	\begin{equation*}
	\begin{aligned}
	\dot{\calV} &= \mathbf\frakp_e^\top \bfM^{-1} \dot{\mathbf\frakp}_e +  \bfe^\top \bfM^{-1} \mathbf\frakp_e + \frac{c_1 \bfe^\top \dot{\mathbf\frakp}_e}{c_2}  + \frac{c_1 \dot{\bfe}^\top \mathbf\frakp_e}{c_2}  + \frac{\bfe_{\bfa}^\top \dot{\bfa}}{c_2}\\
	&= - \mathbf\frakp_e^\top\bfM^{-1}\bfK_{\bfd} \bfM^{-1} \mathbf\frakp_e - \frac{c_1}{c_2}\bfe^\top \bfe\\
	&\qquad  -\frac{c_1}{c_2}\bfe^\top \bfK_{\bfd} \bfM^{-1} \mathbf\frakp_e + \frac{c_1}{c_2} \bfe^\top  \begin{bmatrix} \hat{\bfe}_{\bfomega} & \mathbf{0}\\ \mathbf{0}& \mathbf{0} \end{bmatrix} \mathbf\frakp_e \\
	&\qquad+ \frac{c_1}{c_2} \bfe^\top  \begin{bmatrix} \bfR^\top \bfR^* \hat{\bfomega}^*\bfR^{*\top} \bfR & \mathbf{0}\\ \mathbf{0}& \mathbf{0} \end{bmatrix}  \mathbf\frakp_e \\
	&\qquad+ \frac{c_1}{c_2} \mathbf\frakp_e^\top \bfM^{-1} \begin{bmatrix} k_{\bfp} \bfI & \mathbf{0}\\\mathbf{0} & k_{\bfR}\bfE(\bfR,\bfR^*) \end{bmatrix} \mathbf\frakp_e,
	\end{aligned}
	\end{equation*}
	where we use \eqref{eq:edot}, \eqref{eq:frakpedot}, and that $\bfomega = \bfe_\bfomega  + \bfR^\top \bfR^* \bfomega^*$ by definition of $\bfe_{\bfomega}$. Hence, in the domain $\calT$, we have:
	\begin{equation}
	\frac{d}{dt}\calV \leq - \bfz^\top \bfQ_3 \bfz = - \bfz^\top \begin{bmatrix} q_1 & q_2 \\ q_2 & q_3\end{bmatrix}\bfz,
	\end{equation}
	where $q_1 = \frac{c_1}{c_2}$, $q_2 = -\frac{c_1}{c_2} \prl{ \lambda_{\max}(\bfK_{\bfd}\bfM^{-1}) + \beta + \gamma}$, and 
		$q_3~=~\lambda_{\min}(\bfM^{-1}\bfK_{\bfd}\bfM^{-1}) - \frac{c_1}{c_2} \max\crl{k_{\bfp},k_{\bfR}}\lambda_{\max}(\bfM^{-1})$.
	%
	
	Since $k_{\bfp}$, $k_{\bfR}$, $\bfK_{\bfd} = \diag(k_{\bfv}\bfI,k_{\bfomega}\bfI)$ can be chosen arbitrarily large and $c_1/c_2$ can be chosen arbitrarily small, there exists a choice of constants that ensures that the matrices $\bfQ_1$, $\bfQ_2$, and $\bfQ_3$ are positive definite. Consider the sub-level set of the Lyapunov function $\calR = \{\bfx \in \calT | \calV(\bfx) \leq \delta\}$ where $\delta < \lambda_{\min} (\bfQ_1)\min(\alpha (2-\alpha),\beta)/2$. For $\bfx_0 \in \calR$, we have $d\calV/dt \leq 0$,  $\bfPsi(\bfR, \bfR^*) \leq \frac{\|\bfe_{\bfR}\|_2^2}{2-\alpha} \leq \frac{2\delta}{(2-\alpha)\lambda_{\min}(\bfQ_1)}  < \alpha$, and $\|\bfe_{\bfomega}(\bfx,\bfx^*)\| \leq \frac{2\delta}{\lambda_{\min}(\bfQ_1)} \leq \beta$ for all $\bfx(t), t > 0$, i.e., $\calR$ is a positively invariant set. Therefore, for any system trajectory starting in $\calR$, the tracking errors $\bfe$, $\mathbf\frakp_e$ are asymptotically stable, while the estimation error $\bfe_{\bfa}$ is stable and uniformly bounded, by the LaSalle-Yoshizawa theorem \cite[Thm.~A.8]{krstic1995nonlinear}.
	\qedhere
\end{proof}

%% file: tex/Experiments.tex
\section{Evaluation}
\label{sec:exp_results}

We evaluate our data-driven geometric adaptive controller on a fully-actuated pendulum and under-actuated quadrotor.

\subsection{Pendulum}
\label{subsec:pendulum}
Consider a pendulum with angle $\varphi$, scalar control input $u$, and dynamics $m\ddot{\varphi} = -5\sin{\varphi} + u + d$, where the mass is $m = 1/3$, the potential energy is $V(\varphi) = 5(1-\cos{\varphi})$, the input gain is $g(\varphi) = 1$, and the disturbance $d = - \mu \dot{\varphi}$ models a friction force with unknown friction coefficient $\mu$. To illustrate our geometric adaptive control approach, we consider $\varphi$ as a yaw angle specifying a rotation $\bfR$ around the $z$ axis. The angular velocity is $\bfomega = [0, 0, \dot{\varphi}]$. We remove the position $\bfp$ and linear velocity $\bfv$ terms from Hamilton's equations in \eqref{eq:PH_dyn} to obtain the pendulum dynamics.

\begin{table}[t]
\caption{Tracking errors and disturbance estimation error per time step (mean $\pm$ standard deviation) with our adaptive controller, with disturbance observer (DOB), and without adaptation for $100$ experiments of $10$-second pendulum angle tracking.}
\label{table:pend_tracking_errors}
\centering
\begin{tabular}{ lccc } 
Approach  & No adaptation & Our adaptation & DOB\\
\hline
\hline
Angle error & $0.35 \pm 0.14$ & $\boldsymbol{0.04 \pm 0.02}$ & $0.08 \pm 0.02$ \\
Disturbance error & $0.67 \pm 0.27 $ & $\boldsymbol{0.06 \pm 0.03}$ & $0.10 \pm 0.04$
\end{tabular}
\end{table}

\begin{figure*}[!t]
	\centering
	\begin{subfigure}[t]{0.33\textwidth}
		\centering
		\includegraphics[width=\textwidth]{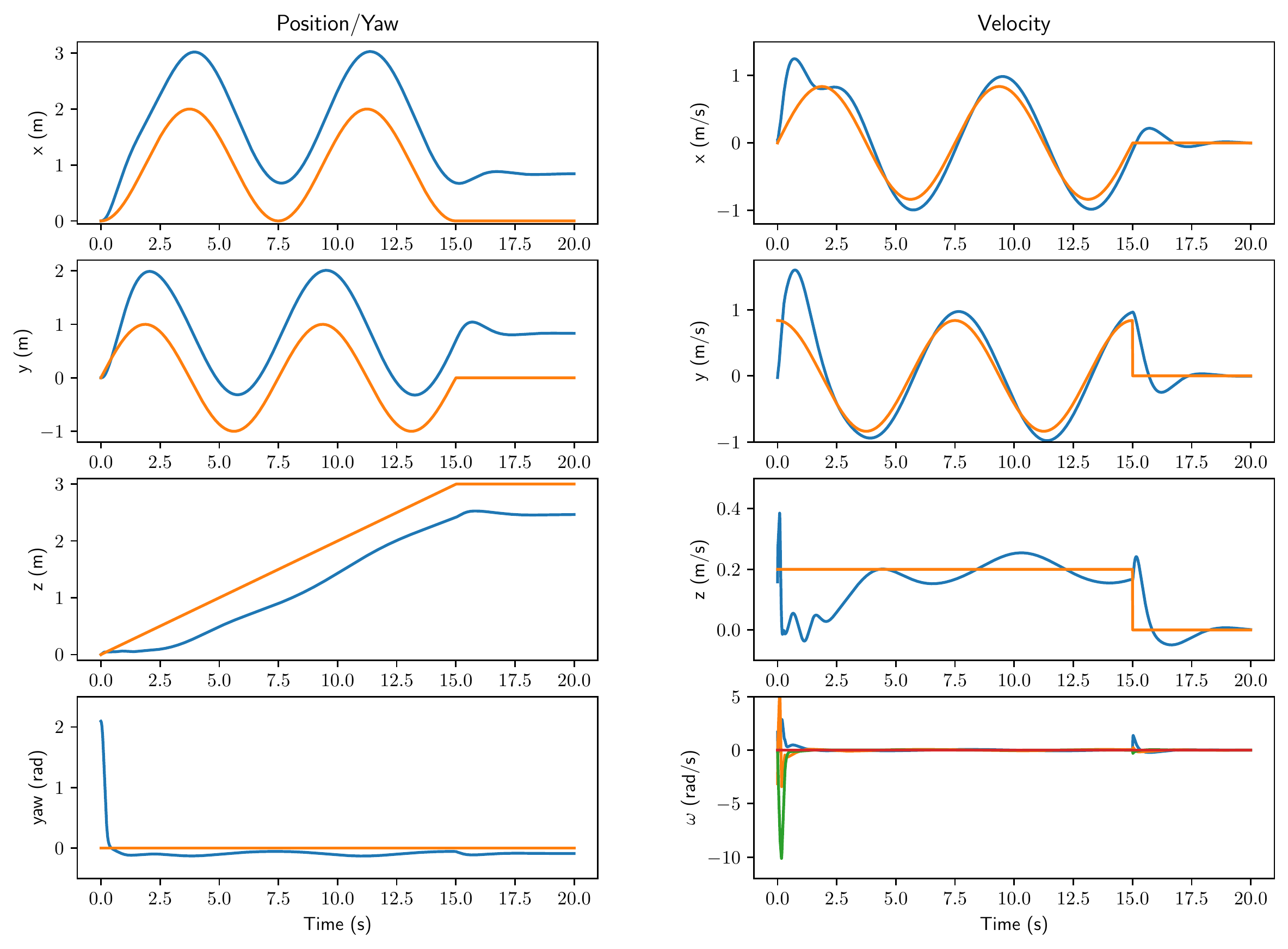}%
		\caption{}
		\label{fig:tracking_results_noadaptive_exp1}
	\end{subfigure}%
	\hfill
	\begin{subfigure}[t]{0.33\textwidth}
		\centering
		\includegraphics[width=\textwidth]{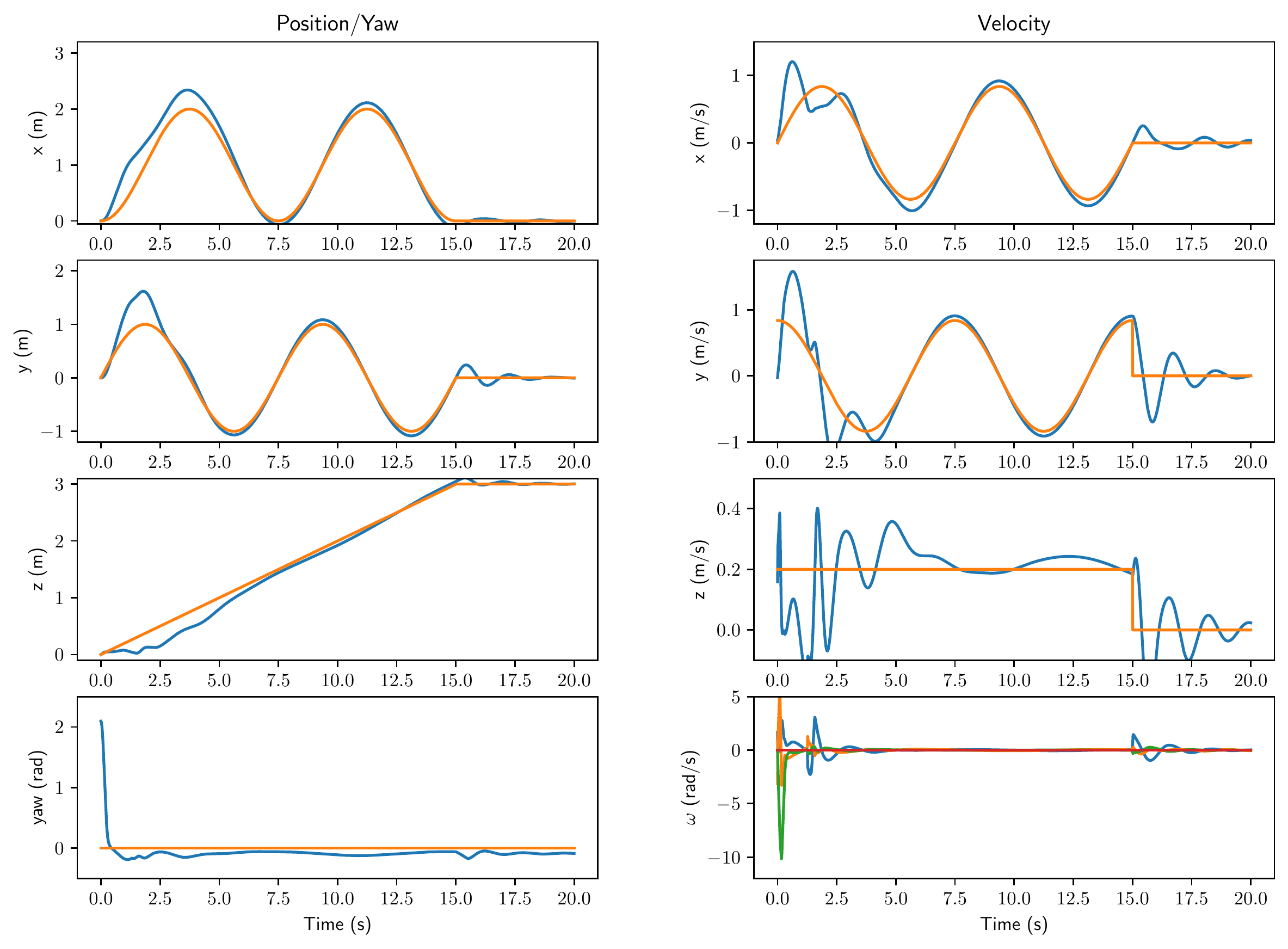}%
		\caption{}
		\label{fig:tracking_results_adaptive_exp1}
	\end{subfigure}%
	\hfill
	\begin{subfigure}[t]{0.33\textwidth}
		\centering
		\includegraphics[width=\textwidth]{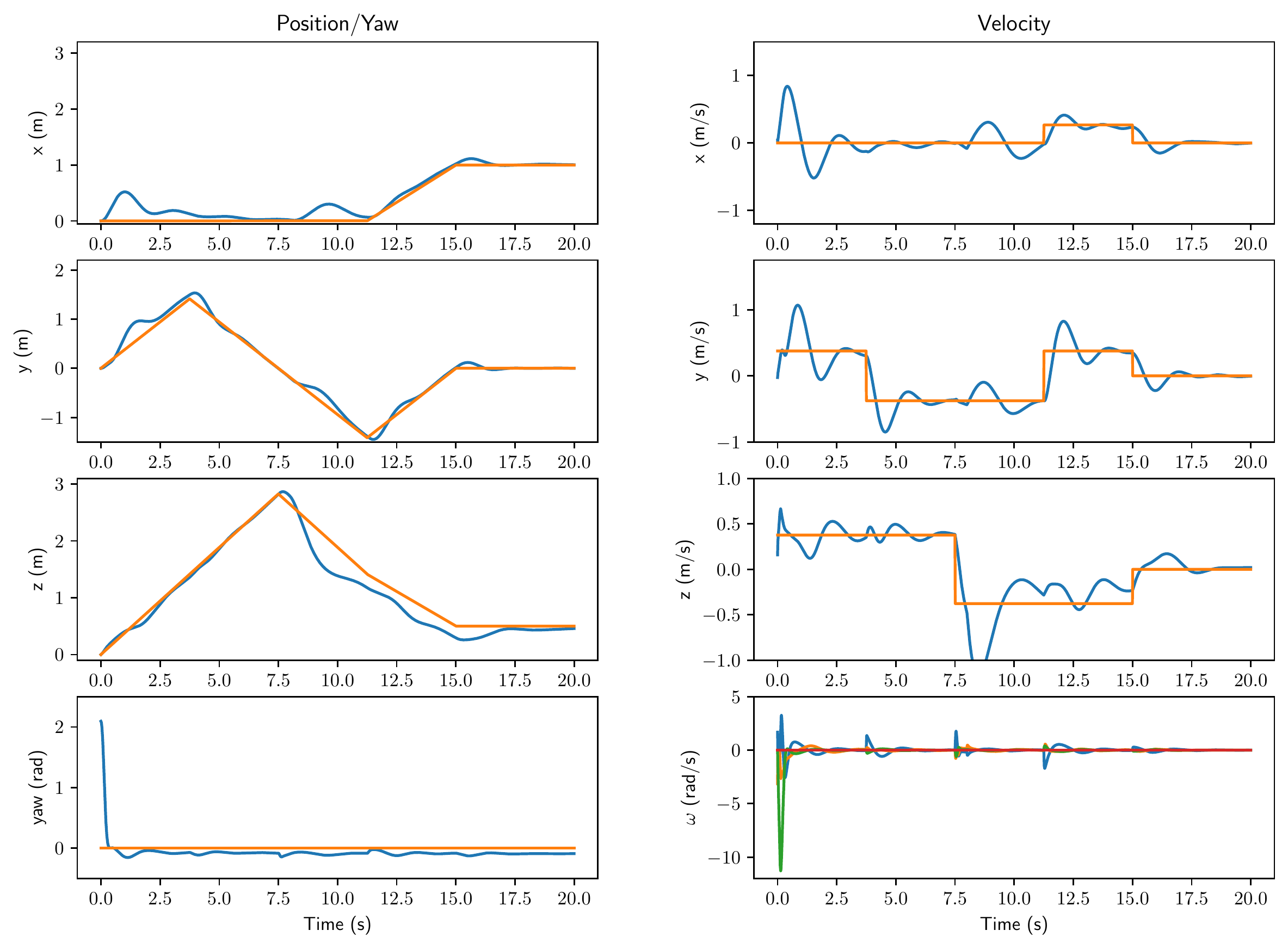}%
		\caption{}
		\label{fig:tracking_results_adaptive_exp2}
	\end{subfigure}
		\caption{Tracking spiral and diamond-shaped trajectories with a PyBullet Crazyflie quadrotor \cite{gym-pybullet-drones2020} under an external wind $\bfd_w =[
			0.075 \quad 0.075 \quad 0
			]$ and two rotors turning defective from the beginning (scenario 1) and after $8$s (scenario 2) both with $(\delta_1,\delta_2)=(80\%, 80\%)$: (a) scenario 1 without adaptation, (b) scenario 1 with adaptation, (c) scenario 2 with adaptation.}
	\label{fig:quadrotor_exp}
\end{figure*}

\begin{figure}[!t]
	\centering
	\includegraphics[width=0.49\linewidth,trim=0mm 0mm 0mm 20mm, clip]{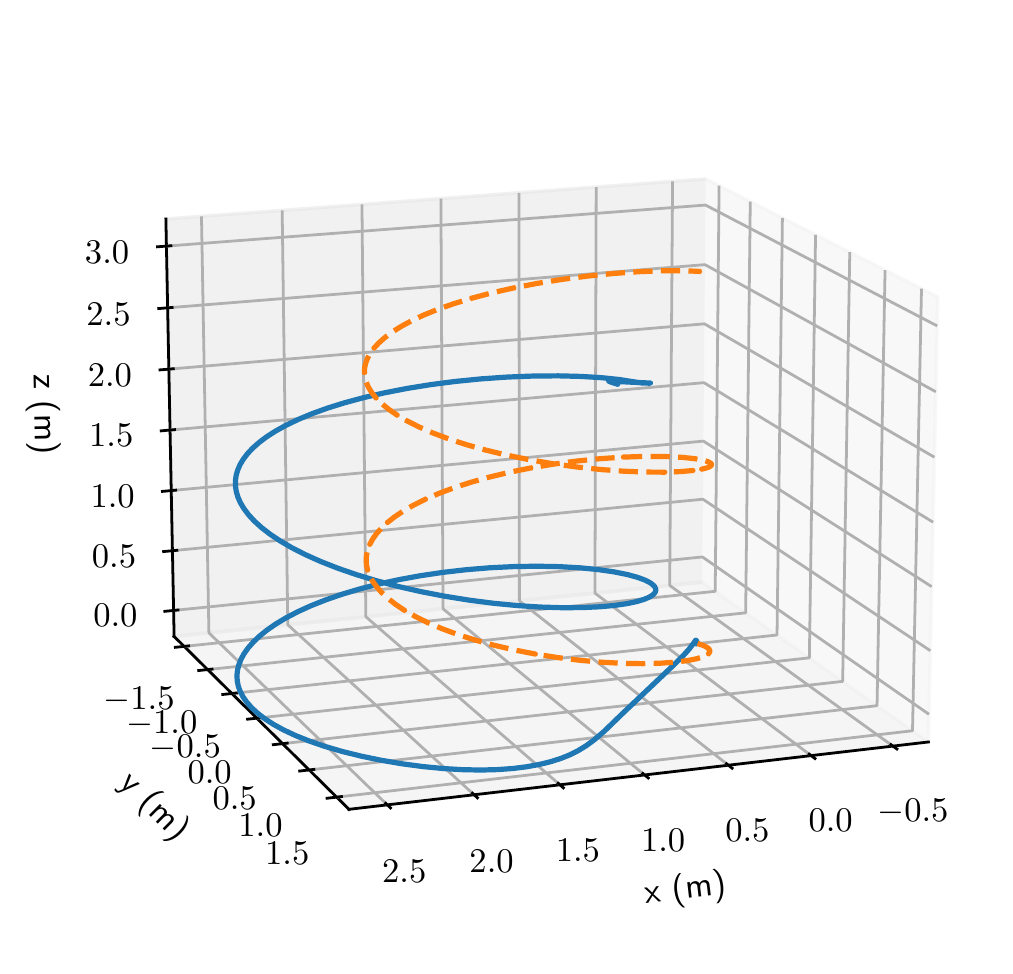}%
	\hfill%
	\includegraphics[width=0.49\linewidth,trim=0mm 0mm 0mm 20mm, clip]{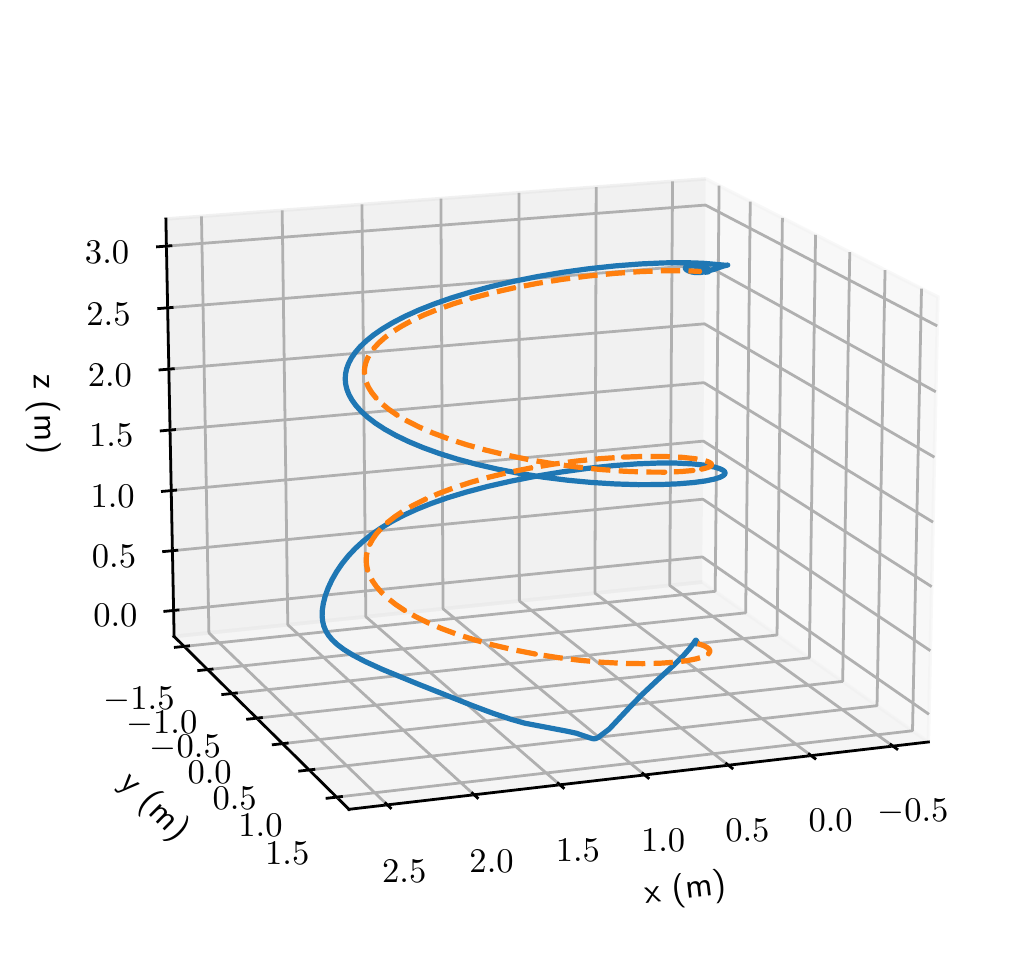}\\%
	\includegraphics[width=0.49\linewidth,trim=0mm 0mm 0mm 12mm, clip]{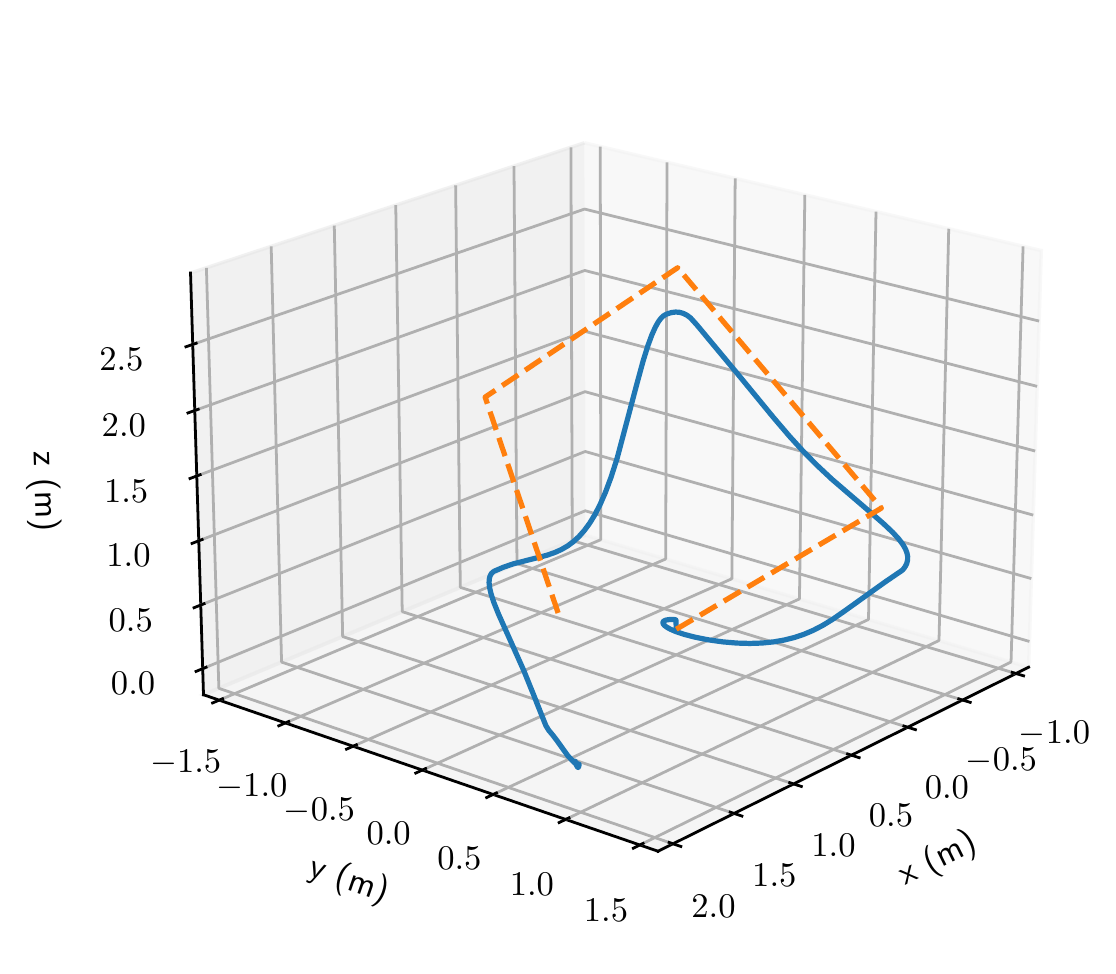}%
	\hfill%
	\includegraphics[width=0.49\linewidth,trim=0mm 0mm 0mm 12mm, clip]{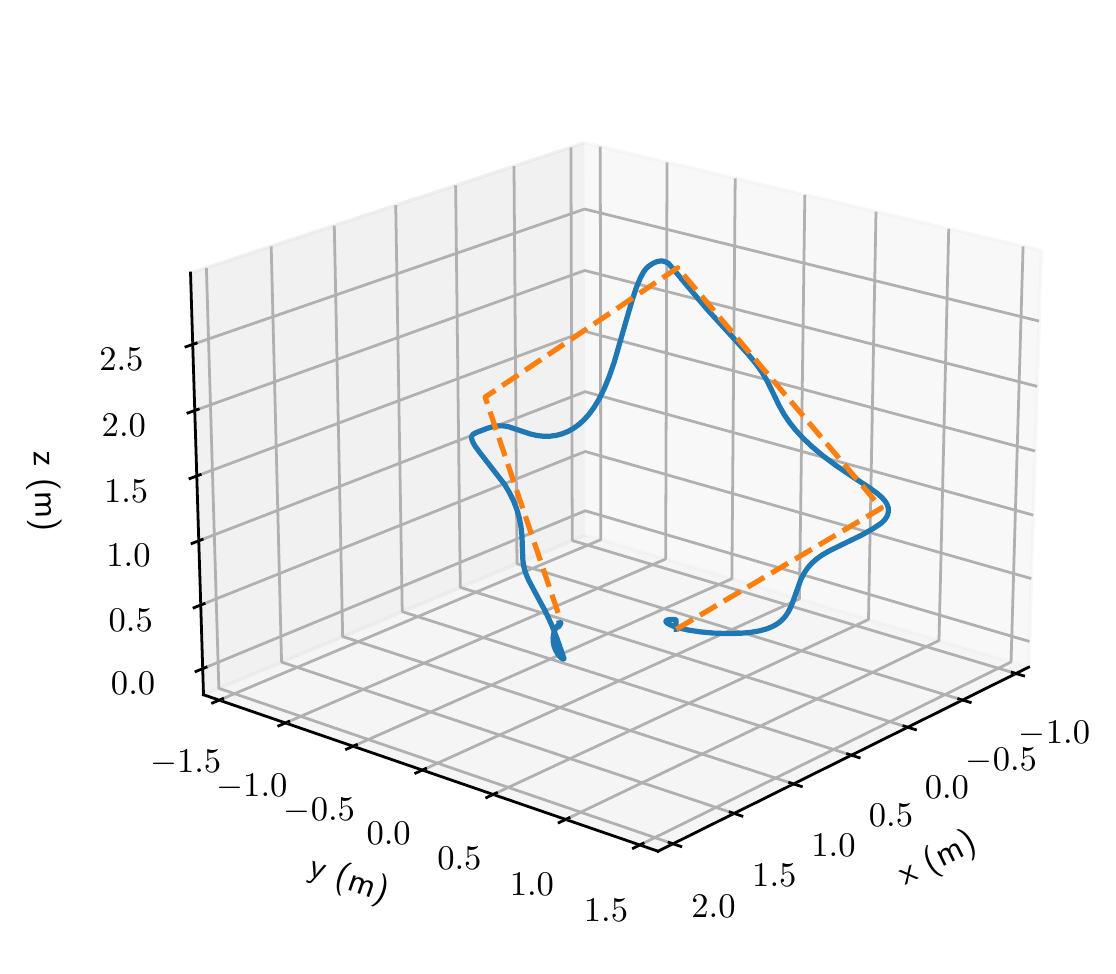}%
	\caption{Tracking visualization for scenario 1 (top) and scenario 2 (bottom) with (right) and without (left) adaptation.}
	\label{fig:quadrotor_exp}
\end{figure}

To learn the disturbance features, we consider $M = 11$ realizations of the disturbance $d_j = -\mu_j \dot{\varphi}$ with friction coefficient $\mu_j = 0.05(j-1) \in [0, 0.5]$ for $j = 1, \ldots, M$. For each value $\mu_j$, we collect transitions $\calD_j = \{\bfx^{(ij)}_{0}, \bfu^{(ij)}, \bfx^{(ij)}_{f}, \tau^{(ij)}\}_{i = 1}^{1024}$ by applying $1024$ random control inputs to the pendulum for a time interval of $\tau^{(ij)}~=~0.01$~s. We train the disturbance model as described in Sec. \ref{subsec:Ham_dyn_learning} for $4000$ iterations with learning rate~$10^{-4}$.

We verify our adaptive controller $(\bfpi,\bfrho)$ in Sec. \ref{subsec:adaptive_control_online} with the task of tracking a desired angle $\varphi^*(t) = \pi t/5 + \pi t^2/50$. We simplify the controller $\bfpi$ in  \eqref{eq:ES_DI_COMP_control} and the adaptation law $\bfrho$ in \eqref{eq:geometric_adaptive_law} by removing the position and linear velocity components. The controller gains are: $k_{\bfR} = 1$, $k_\bfd = 2$, $c_{\bfR} = 75$, $c_{\bfomega} = 10$.
We compare our approach with a disturbance observer method \cite{chen2000nonlinear, li2014disturbance} for the pendulum. As the disturbance features in \eqref{eq:f_external} are unknown, we design an observer to estimate $d$ online. Let $z$ be the observer state with dynamics $m\dot{z} = -l(\dot{\varphi})z - l(\dot{\varphi})\prl{r(\dot{\varphi}) - 5 \sin(\varphi) + u}$,
where $l(\dot{\varphi}) = \frac{\partial r(\dot{\varphi})}{\partial \dot{\varphi}}$ for some function $r(\dot{\varphi})$. The disturbance $d$ is estimated as $\hat{d} = z + r(\dot{\varphi})$.
The disturbance estimation error is $e_d = \hat{d} - d$ satisfying $\dot{e}_d = \dot{z} + \frac{\partial r(\dot{\varphi})}{\partial \dot{\varphi}} \ddot{\varphi} = -l(\dot{\varphi})e_d/m$.
We choose $r(\dot{\varphi}) = \dot{\varphi}$ so that the disturbance estimation errors converges to $0$ asymptotically. The estimated disturbance is compensated using the same tracking controller $\bfpi$ in  \eqref{eq:ES_DI_COMP_control}.

We run the experiments $100$ times with a friction coefficient $\mu$ uniformly sampled from the range $[0.5, 3]$. Table \ref{table:pend_tracking_errors} shows the angle tracking errors and the disturbance estimation errors with our adaptive controller, with the disturbance observer (DOB), and without adaptation. Our adaptive controller achieves better tracking error and disturbance estimation error than the DOB approach. Fig.~\ref{fig:pend_exp} plots the tracking errors and disturbance estimation error with $\mu = 2.5$, showing that we achieve the desired angle $\varphi^*(t)$ and are able to converge to the state-dependent ground-truth disturbance $d_{gt} = -2.5\dot{\varphi}$. Without knowing the disturbance features, the DOB method lags behind the changes in ground-truth disturbances caused by the velocity $\dot{\varphi}$. This illustrates the benefit of our approach -- the learned disturbance features improve the performance of the adaptive controller.

\begin{table}[t]
\caption{Tracking errors per time step (mean $\pm$ standard deviation) for $100$ experiments of quadrotor trajectory tracking.} 
\label{table:tracking_errors}
\centering
\begin{tabular}{ lcc } 
Experiments  & Diamond-shaped & Spiral\\
\hline
\hline
Scenario 1 (without adaptation) & $0.71 \pm 0.15$ & $0.55 \pm 0.13$ \\
Scenario 1 (with adaptation) & $\boldsymbol{0.12 \pm 0.02}$ & $\boldsymbol{0.13 \pm 0.01}$ \\
Scenario 2 (without adaptation) & $0.62 \pm 0.13$ & $0.64 \pm 0.10$ \\
Scenario 2 (with adaptation) & $\boldsymbol{0.12 \pm 0.02}$ & $\boldsymbol{0.16 \pm 0.02}$ \\
\end{tabular}
\end{table}

\subsection{Crazyflie Quadrotor}


Next, we consider a Crazyflie quadrotor, simulated using the PyBullet physics engine \cite{gym-pybullet-drones2020}, with control input $\bfu = [f, \bftau]$ including the thrust $f\in \mathbb{R}_{\geq 0}$ and torque $\bftau \in \mathbb{R}^3$ generated by the $4$ rotors. The mass of the quadrotor is $m = 0.027$ kg and the inertia matrix is $\bfJ = 10^{-5}\diag([1.4, 1.4, 2.2])$, leading to the generalized mass matrix $\bfM(\frakq) = \diag(m\bfI, \bfJ)$. The potential energy is $V(\frakq) = mg\begin{bmatrix}
0 & 0 & 1
\end{bmatrix} \bfp$, where $\bfp$ is the position of the quadrotor and $g\approx 9.8\;ms^{-2}$ is the gravitational acceleration. We consider disturbances from three sources: 1) horizontal wind, simulated as an external force $\bfd_w = \begin{bmatrix}
w_{x} & w_{y} & 0
\end{bmatrix}^\top \in \bbR^3$ in the world frame, i.e., $\bfR^\top\bfd_w$ in the body frame; 2) two defective rotors $1$ and $2$, generating $\delta_1$ and $\delta_2$ percents of the nominal thrust, respectively; and 3) near-ground, drag, and downwash effects in the PyBullet simulated quadrotor. 

As described in Sec. \ref{subsec:Ham_dyn_learning}, we learn the disturbance features $\bfW_\bftheta(\frakq, \frakp)$ from a dataset $\calD$ of transitions using a Hamiltonian-based neural ODE network. We collect a dataset $\calD = \{\calD_j\}_{j = 1}^M$ with $M = 8$ realizations of the disturbances $\bfd_{wj}$, $\delta_{1j}$, and $\delta_{2j}$. Specifically, the wind components $w_{xj}, w_{yj}$ are chosen from the set $\{\pm 0.025, \pm 0.05\}$, while the values of $\delta_{1j}$ and $\delta_{2j}$ are sampled from the range $[94\%, 98\%]$. For each disturbance realization, a PID controller provided by \cite{gym-pybullet-drones2020} is used to drive the~quadrotor~from~a~random starting point to $9$ different desired poses, providing transitions  $\calD_j~=~\{\bfx^{(ij)}_{0}, \bfu^{(ij)}, \bfx^{(ij)}_{f}, \tau^{(ij)}\}_{i = 1}^{1080}$ with $\tau^{(ij)}~=~1/240$~s.

We verify our geometric adaptive controller with learned disturbance features by having the quadrotor track pre-defined trajectories in the presence of the aforementioned disturbances $\bfd$. The desired trajectory is specified by the desired position $\bfp^*(t)$ and the desired heading $\bfpsi^*(t)$. 
We construct an appropriate choice of $\bfR^*$ and $\bfomega^*$ from $\bfpsi^*(t)$, as described in \cite{duong21hamiltonian, goodarzi2015geometric}, to be used with the adaptive controller. 
The tracking controller in \eqref{eq:ES_DI_COMP_control} with gains $k_\bfp = 0.135, k_\bfv = 0.0675, k_{\bfR} = 1.0,$ and $ k_{\bfomega} = 0.08$, is used to obtain the control input $\bfu$ that compensates for the disturbances. The disturbances $\bfd$ are estimated by updating the weights $\bfa$ according to the adaptation law \eqref{eq:geometric_adaptive_law} with gains $c_{\bfp}= c_{\bfv} = c_{\bfR} = c_{\bfomega} = 0.04$.

We test the controller with wind $\bfd_w$, rotors $1$ and $2$ that become defective from the beginning (scenario 1) or during flight at $t=8$ s (scenario 2), and near-ground, drag, and downwash effects enabled in PyBullet. We track diamond-shaped and spiral trajectories $100$ times with $w_x$ and $w_y$ uniformly sampled from $[0, 0.075]$ and $\delta_1$ and $\delta_2$ drawn uniformly from $[80\%, 99\%]$. Table~\ref{table:tracking_errors} shows the mean and standard deviation of the tracking errors with and without adaptation from the $100$ flights. The errors with adaptation are $\sim 5$ times lower than without adaptation, illustrating the benefit of our adaptive control design.
For $\bfd_w = \begin{bmatrix}
 0.075 & 0.075 & 0
 \end{bmatrix}$ and $(\delta_1,\delta_2)=(80\%, 80\%)$, the quadrotor in scenario 1 without adaptation drifts as seen in Fig. \ref{fig:tracking_results_noadaptive_exp1} and \ref{fig:quadrotor_exp} (upper-left) while our adaptive controller estimates the disturbances online after a few seconds and successfully tracks the trajectory as seen in Fig. \ref{fig:tracking_results_adaptive_exp1} and \ref{fig:quadrotor_exp} (upper-right). For the same disturbance, the quadrotor in scenario 2 with our controller starts to track the trajectory, then drops down at $t=8$ s, due to the rotors becoming defective, but recovers as our adaptation law updates the disturbances accordingly, as seen in Fig. \ref{fig:tracking_results_adaptive_exp2} and \ref{fig:quadrotor_exp} (lower-right). Without adaptation, the quadrotor crashes to the ground at $t\approx 12.5$s, shown in Fig. \ref{fig:quadrotor_exp} (lower-left).

%% file: tex/Conclusion.tex
\section{Conclusion}
\label{sec:conclusion}

This paper introduced a neural ODE network for disturbance feature learning using disturbance-corrupted trajectory data from a rigid-body system with Hamiltonian dynamics. To enable trajectory tracking with online disturbance compensation, we designed a passivity-based tracking controller and augmented it with an adaptation law that compensates disturbances relying on the learned features and geometric tracking errors. Our evaluation showed that our geometric adaptive controller quickly estimates disturbances online and successfully tracks desired trajectories, outperforming adaptation methods that employ online disturbance estimation without learned disturbance features. Future work will focus on deploying the geometric adaptive controller on real UGV and UAV robot systems.